\documentclass[accepted]{uai2021} 
\usepackage[american]{babel}

\usepackage{natbib} 
\usepackage{mathtools} 
\usepackage{booktabs} 
\usepackage{tikz} 
\usepackage{dsfont}
\usepackage{amsmath}
\DeclareMathOperator*{\argmax}{argmax}
\DeclareMathOperator*{\argmin}{argmin}

\title{Explicit Pairwise Factorized Graph Neural Network \\ for Semi-Supervised Node Classification}

\author[1,2]{\href{mailto:Yu Wang <yu1024.wang@tum.de>?Subject=Your UAI 2021 paper}{Yu Wang}{}}
\author[1]{Yuesong Shen}
\author[1]{Daniel Cremers}
\affil[1]{%
    Technical University of Munich, Germany
}
\affil[2]{%
    University of Illinois at Chicago, USA
}

\begin{document}
\maketitle

\begin{abstract}
Node features and structural information of a graph are both crucial for semi-supervised node classification problems. A variety of graph neural network (GNN) based approaches have been proposed to tackle these problems, which typically determine output labels through feature aggregation. This can be problematic, as it implies conditional independence of output nodes given hidden representations, despite their direct connections in the graph. To learn the direct influence among output nodes in a graph, we propose the \textbf{E}xplicit \textbf{P}airwise \textbf{F}actorized \textbf{G}raph \textbf{N}eural \textbf{N}etwork (EPFGNN), which models the whole graph as a partially observed Markov Random Field. It contains explicit pairwise factors to model output--output relations and uses a GNN backbone to model input--output relations. To balance model complexity and expressivity, the pairwise factors have a shared component and a separate scaling coefficient for each edge. We apply the EM algorithm to train our model, and utilize a star-shaped piecewise likelihood for the tractable surrogate objective. We conduct experiments on various datasets, which shows that our model can effectively improve the performance for semi-supervised node classification on graphs.
\end{abstract}

\section{Introduction}

 We live in a world full of interconnections. For example, publications are connected with citation links; Social media users are connected when they follow each other \citep{qu2019gmnn}; protein structures are interconnections of amino acids. Modeling relational data is an important topic for machine learning. 
 These relational data can be represented by graphs, which model a set of objects with node features and their relationships with graph edges \citep{zhou2018graph}.

 A large number of systems, including social networks \citep{hamilton2017inductive}, citation networks \citep{kipf2016semi}, physical systems \citep{sanchez2018graphphysics}, protein interactions \citep{fout2017graphprotein}, knowledge graphs \citep{hamaguchi2017graphknowledge}, etc.\ can be represented as graphs. And graph-based learning is receiving increasing attention from researchers.

In this paper, we focus on the problem of classifying nodes of a graph, such as a graph of a citation network. This problem can be framed as graph-based semi-supervised learning since the label information is available for only a small subset of nodes.

For problems involving data which have grid-like graph structures, e.g., images and videos, Convolutional Neural Networks (CNN) have achieved state-of-the-art results \citep{velivckovic2017graph}. However, many problems involve data such as 3D meshes and citation networks, which are represented by graphs having a general structure. An increasing number of researchers try to generalize the notion of convolution for general graph structure, motivated by its success in the domain of computer vision, to address the problems involving general graphs. And there are many attempts to extend neural networks to deal with graphs with arbitrary structure: Graph Neural Networks (GNNs), such as Graph Convolutional Networks (GCN) \citep{kipf2016semi} and Graph Attention Networks (GAT) \citep{velivckovic2017graph} have been receiving increasing attention because of their effective node representation learning on general graphs. These methods take into account the graph structure and aggregate features from neighbor nodes.

However, one weakness of these methods is that they neglect direct local dependency among output label nodes, which can be important for node classification. To learn the interdependency of connected node labels, we propose to model the whole graph with a pairwise partially observed Markov Random Field (MRF) and use explicit pairwise factors to model direct dependencies between connected node labels. To benefit from the representational power of GNNs, we apply a GNN backbone to learn appropriate unary factors that can reflect the influence from the node input features on node output labels. To avoid overfitting, we use a shared compatibility matrix to parametrize the pairwise factors, which is further multiplied by an edgewise coefficient to reflect the difference among the edges in a graph. 

The partially observed MRF representation learns node representation and node label interdependency simultaneously, taking into account both observed and unobserved nodes. However, the conventional learning objective of maximizing observed data log-likelihood requires marginalizing unobserved nodes, which has an exponential computational cost and is thus intractable. To address this challenge, we use the EM algorithm \citep{koller2009pgmbook} to learn the MRF and maximize instead the evidence lower bound with alternating E-steps and M-steps: in each E-step we update the candidate distribution on the unobserved nodes, which is then used in the subsequent M-step to maximize the expected complete data log-likelihood.

This alternating training is still challenging because the inference process on general loopy graphs is known to be intractable \citep{koller2009pgmbook}. To remedy this issue and perform training efficiently, we introduce a piecewise likelihood \citep{sutton2009piecewise} based local training method, which provides a surrogate objective that makes the inference process more manageable. It divides the graph into tractable subgraphs for easy inference, and avoids the intractable global inference procedure.

Our contributions are:
\begin{itemize}

\item We propose the explicit pairwise factorized graph neural network (EPFGNN), which uses explicit pairwise output--output factors to augment a graph neural network backbone so that we can model direct dependencies among output label nodes in graphs.

\item We use a shared compatibility matrix to parametrize pairwise factors and extend it with an edgewise multiplicative scaling coefficient. This ensures that the model can aggregate the neighbor node features and label information flexibly on both assortative graphs and disassortative graphs, while simultaneously avoid overfitting. 

\item We propose an EM-based learning procedure, and employ piecewise likelihood as surrogate objective to make the training on MRF tractable while retaining the structural information.

\item We conduct a series of experiments \footnote{\url{https://github.com/YuWang-1024/EPFGNN}}, which shows that our model, along with the proposed training procedure, can effectively tackle semi-supervised node classification on graphs, outperforming existing baselines.
\end{itemize}

\section{Related Work}\label{sec::related work}

\paragraph{Graph neural network}
Motivated by the effectiveness of CNNs, an increasing number of researchers aim to design convolution operation on graphs to extract information from connected neighbor node features. The resulting graph neural networks are generally defined by a message-passing scheme as follows:
\begin{equation}
    h_i^{k+1} = update(h_i^k, aggregate_{j\in \mathcal{N}(i)}(h_i^k, h_j^k) ), 
\end{equation}
where $h_i^k$ is the hidden representation of a node at k-th layer, $\mathcal{N}(i)$ is the set of neighbor nodes connected with node $i$. The node representations are updated by weighted aggregations of the features from each central node and its neighbors. A variety of aggregation and update functions have been proposed by different GNN variants.

An effective GNN model named Graph Convolution Network (GCN) by \citet{kipf2016semi} reduces the computational complexity of eigen-decomposition involved in spectral graph convolution by introducing several approximations. The GCN layer is defined as follows:
\begin{equation}
    H^{k+1} = \sigma (\tilde{D}^{-\frac{1}{2}}\tilde{A}\tilde{D}^{-\frac{1}{2}}H^{k} W^k),
\end{equation} 
where $\tilde{A} = A + I_N$ is the adjacency matrix with self-connections, and $\tilde{D}_{ii} = \sum_j \tilde{A}_{ij}$ indicates the degree of node $i$.
Since the graph has an arbitrary structure, the number of neighbors varies for different nodes, and can be large in graphs of massive scale. It is then inefficient to aggregate features from all neighbor nodes. To address this issue, GraphSAGE \citep{huang2018adaptive} applies a layer-wise sampler to control the size of neighbors. While GraphSAGE assumes the contribution of neighbors to the central node has equal importance, the method named Graph Attention Network (GAT) \citep{velivckovic2017graph} applies an attention mechanism to weigh the different contributions from neighbor node features to the central node.

While the aforementioned GNN models can effectively process graph structured data for semi-supervised node classification, 
they neglect the direct interdependency among output nodes despite their connections in the graph structures. To address this problem, graph Markov neural network (GMNN) \citep{qu2019gmnn} proposes a pseudo-likelihood based variational EM framework to model the interdependency among node labels. In the M-step, they optimize the graph neural network by maximizing the pseudo-likelihood. The pseudo-likelihood assumes the conditioning of connected node labels This assumption deviates from typical semi-supervised settings since many of the nodes are unobserved. In E-steps, GMNN uses another graph neural network to aggregate the information from connected node labels. This implicitly assumes that the graph is assortative, as typical feature aggregation operation in GNNs perform smoothing. However, in case of disassortative graphs where connected nodes tend to disagree with each other, the above assumption no longer holds. 

\paragraph{Surrogate objective function}
Given a graph representation $\mathcal{G} = (V, E)$, where $V$ is the set of nodes, and $E$ is the set of edges. Denote the set of node features as $\mathcal{X}$ and the corresponding node labels as $\mathcal{Y}$. We employ a Markov random field \citep{lafferty2001conditional} \citep{koller2007introduction} to model the node classification problem. Since the inference on the MRF with loops is known to be intractable \citep{koller2009pgmbook}, the common learning strategy of maximizing data likelihood is infeasible \citep{nowozin2011structured}. This issue motivates the finding of a surrogate objective function $p_{approx}(y|x)$ that approximates the original likelihood. There are two kinds of surrogate training objectives named pseudo-likelihood (PL) \citep{koller2009pgmbook} and piecewise training (PW) objective \citep{sutton2009piecewise}.

Pseudo-likelihood ($p_{PL}$) consists of the product of nodewise conditional likelihood given its Markov blanket. The intuition is that we can obtain the probability density of a node if its Markov blanket is observed. However, the pseudo-likelihood is known to have difficulties modeling longer-range dependencies \citep{koller2009pgmbook}. 
\begin{equation}
    p_{PL}(y|x) = \prod_{n \in V} p_{PL}(y_n|y_{\mathcal{N}(n)},x).
\end{equation}

Piecewise training objective ($p_{PW}$) \citep{sutton2009piecewise} is an alternative way to make the learning on MRF tractable. It models the graph using a factor graph and split the whole graph into tractable subgraphs called pieces. It is inspired by the intuition that the global distribution will be reasonable if all the local factors fit the data well. Given the set of subgraphs $P$, the set of the factors $F$ within each subgraph $R \in P$, and the partition function $Z_R(x)$ of each subgraph that normalizes the piecewise likelihood, the piecewise training objective can be formulated as
\begin{equation}
    p_{PW}(y|x) = \prod_{R \in P} \frac{1}{Z_R(x)}\prod_{F \in R} \phi_{F}(y_F,x).
\end{equation}
Piecewise training objective typically yields better results compared to pseudo-likelihood \citep{sutton2009piecewise}, as the factors in the original likelihood are conserved in the piecewise training objective, and the interdependencies among nodes are better retained.

\section{Methodology}\label{sec:methodology}

We denote a graph as $\mathcal{G} = (V, E)$ , where $V$ is the set of nodes, $E$ is the set of edges. Under the framework of semi-supervised node classification, the full node features $x_V$ and a small subset of node labels $Y_V$ are observed in the graph. Thus, node labels are split into labeled and unlabeled subset $Y_V = (Y_L, Y_U)$. Our goal is to learn a model that can predict these unknown classes $Y_U$ based on the full input features $x_V$, observed output labels $y_L$, and the graph structure.

\subsection{Model representation and learning objective}

\begin{figure}[t]
    \centering  
    \includegraphics[width=0.8\linewidth]{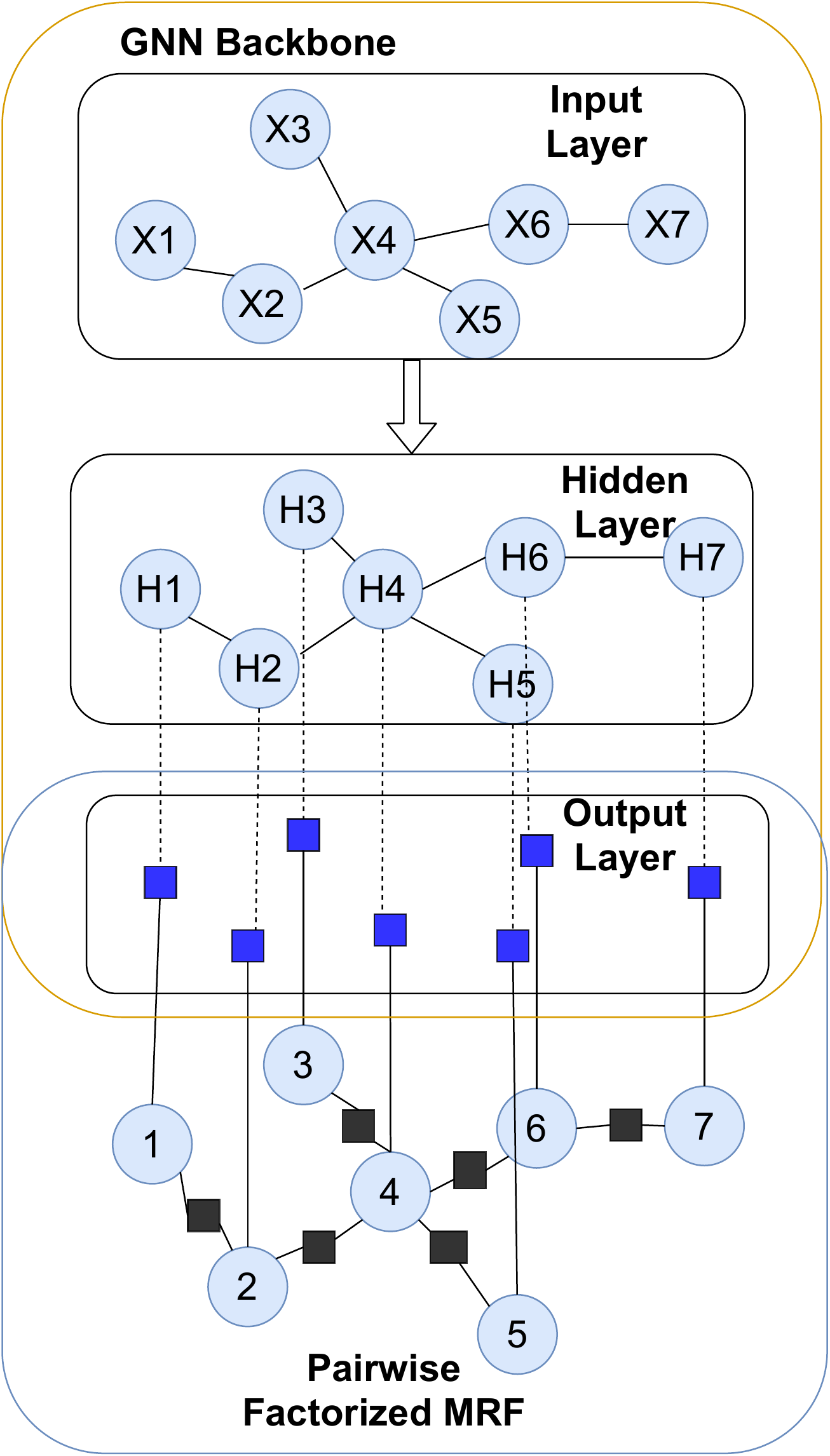}
    \caption{Model representation of EPFGNN. We use the GNN backbone to learn the valid input--output factors that reflect the influence of input node features on the node output labels. The output of GNN backbone is used as the input--output factor value. Furthermore,  every connected node pairs are connected with a pairwise factor. We use the shared compatibility matrix of size $c\times c$ to parametrize the pairwise factor, where $c$ is the output dimension.}
    \label{fig:model_representation}
\end{figure}

\begin{figure*} 
    \centering
    \includegraphics[width=\textwidth]{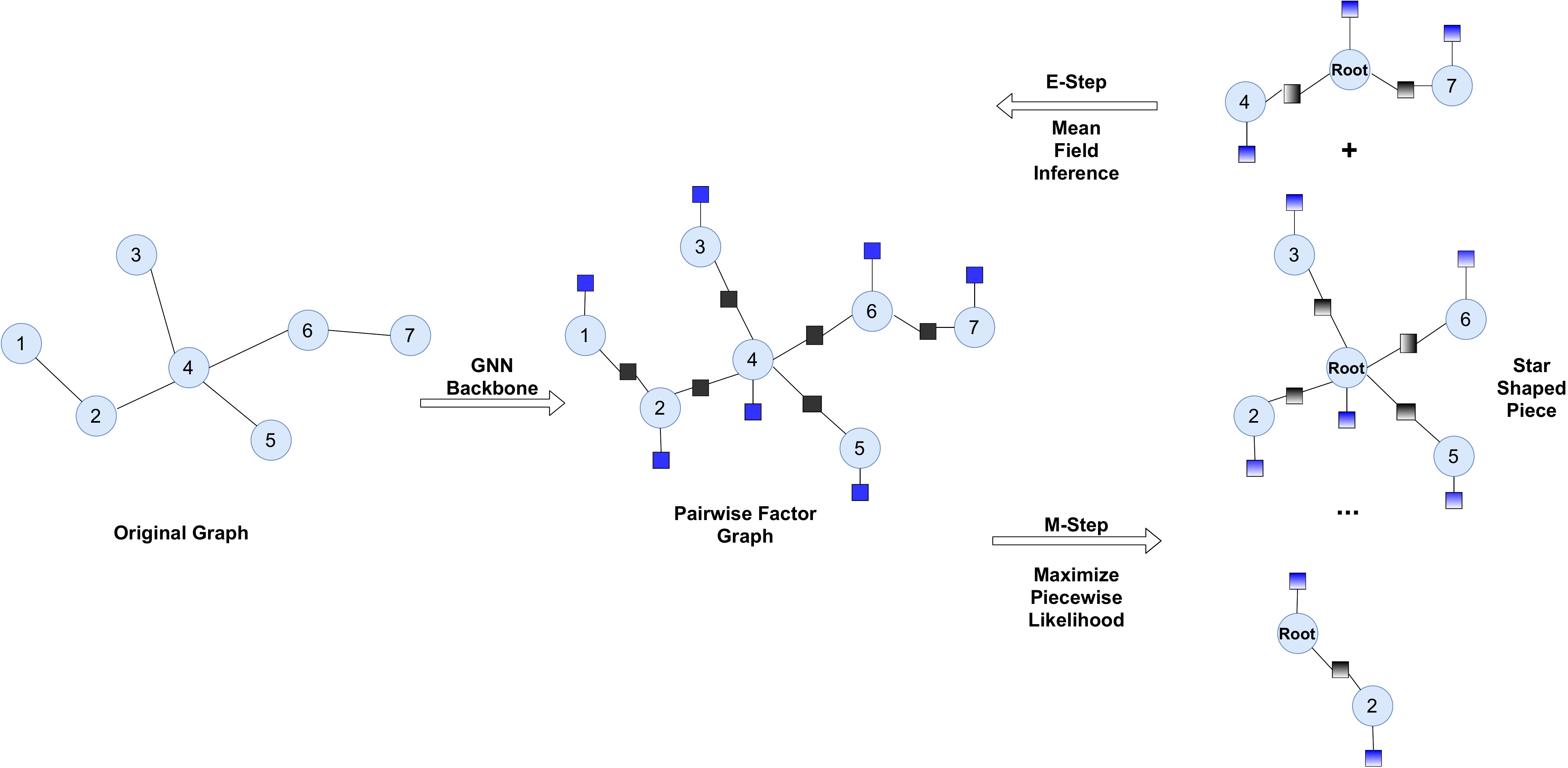} 
    \caption{EPFGNN training framework. The GNN backbone output is used as input--output factor value.  We additionally parametrized the pairwise factor with a compatibility matrix for every connected node pair. We apply EM alternating training method to tackle the intractable challenge caused by partial observation on MRF. In the M-step, we split the graph into star-shaped pieces and maximize the expected complete log-likelihood for every piece. To avoid overcounting, we redistribute the factor values that appear in more than one piece. In the E-step, we merge the pieces back to the original MRF and estimate the posterior distribution $q$ with fixed factor values using the mean field inference method.}
    \label{fig: PWEM}
\end{figure*}

To achieve our goal, we propose the explicit pairwise factorized graph neural network (EPFGNN). We represent the overall distribution $P(Y_V|x_V)$ by a MRF and assume that it admits the following factor decomposition:
\begin{equation}\label{eq:crffactor}
\begin{aligned}
    &P(Y_V|x_V) \\
   &=\frac{1}{Z(x_V)} \prod_{i \in V} \phi_i(Y_i, x_V) \prod_{(j,k) \in E, j < k} \phi_{j,k}(Y_j, Y_k).
\end{aligned}
\end{equation}

In the above expression, all factors are constrained to be positive, and $Z(x_V)$ is the partition function that ensures normalization of the distribution. As shown in Figure~\ref{fig:model_representation}, this factorization needs two kinds of factors to model the whole distribution:
\begin{itemize}
    \item The input--output factors $\phi_i(Y_i, x_V)$ represent the influence of input node features $x_V$ on output node $Y_i$. These are high order factors since they involve all input features, which make direct factor parametrization intractable. To solve this issue, we use a single graph neural network (GNN) backbone to parametrize all these factors so their values can be obtained via feed-forward inference: given an input $x_v$, the GNN backbone should produce a prediction of $\phi_i(y_i, x_v)$ for each output node $i \in V$ and possible output label $y_i$.
    \item The output--output factors $\phi_{j,k}(Y_j, Y_k)$ represent the influence between connected node labels. They are pairwise factors and model the direct dependency between labels for every connected node pair. To avoid overfitting, we parametrize these factors with a single shared symmetric compatibility matrix $K$ of size $c \times c$, where $c$ is the output dimension.
\end{itemize}

Thus our EPFGNN model is parametrized by parameters $\theta$ of the GNN backbone and a compatibility matrix $K$. Our learning objective is to find the parameters $\theta^*$ and $K^*$ that maximize the observed data log-likelihood of labeled output $\tilde{y}_L$ given the input features $\tilde{x}_V$:
\begin{equation}
    \theta^*, K^* = \argmax_{\theta, K} \log P(Y_L = \tilde{y}_L | \tilde{x}_V; \theta, K).
\end{equation}

\subsection{Piecewise EM training framework}

Direct MRF learning under semi-supervised setting is intractable because of the following two challenges:
\paragraph{Partial observation}
Since the output subset $Y_U$ has no observation, we are in the partial observation case where $Y_U$ are unobserved variables. There are two approaches \citep{koller2009pgmbook} to tackle this problem in general: We can either perform gradient ascent to maximize the observed data likelihood directly or use the expectation-maximization (EM) algorithm. In our case, computing the observed log-likelihood directly is computationally infeasible, since we need to marginalize the unobserved random variables $Y_U$, and complex graph structure makes the marginalization computation grows exponentially. Therefore, we apply the EM algorithm, which maximizes instead the expected complete data likelihood w.r.t.\ estimated posterior distribution. We will discuss the details in Section~\ref{sec: EM}.

\paragraph{Intractable inference}
It is well known that the inference process on a general MRF is intractable \citep{koller2009pgmbook} because of the complex graph structure. This makes the learning process impractical beyond simplistic settings. One approach to tackle this problem is to use approximate probabilistic inference methods. However, they are commonly iterative procedures, which are unfriendly for auto-differentiation and hard to ensure convergence. We apply an alternative approach to replace the log-likelihood objective with a more manageable surrogate objective, which makes it easy to do inference and learning. The surrogate objective we use is the piecewise training objective \citep{sutton2009piecewise}. We will elaborate the details in Section~\ref{sec: pw m-step}.

\subsubsection{EM framework}\label{sec: EM}

In the semi-supervised node classification setting, many nodes are unlabeled, which means factor values of the complete data likelihood (Equation~\eqref{eq:crffactor}) are undetermined. As the direct maximization over the observed log-likelihood $\log P(y_L | x_V)$ involves intractable marginalization of unobserved output nodes $Y_U$, we apply the EM framework to leverage the information from unlabeled nodes. EM algorithm maximizes a lower bound of the observed log-likelihood: it finds a variational distribution denoted as $q(Y_U|x_V)$ to approximate the posterior distribution of unlabeled nodes $p(Y_U|y_L,x_V;\theta, K)$, and maximizes the expected complete log-likelihood over all nodes w.r.t.\ the distribution of $q(Y_U|x_V)$. This lower bound comes from our wish to fill in the missing values $y_U$ with a proposal distribution $q(Y_U|x_V)$ and employ the (expected) complete log-likelihood for optimization instead. So the objective function is changed as follows, which is referred to as the evidence of lower bound (ELBO):
\begin{equation} \label{eq:elbodef}
\begin{aligned}
   \mathcal{L}(q; \theta, K) = &\mathds{E}_{q(Y_U|x_V)}[\log P(Y_U,y_L | x_V;\theta, K)]\\
    &+ \sum_{Y_U} - q(Y_U|x_V) \log q(Y_U|x_V). 
\end{aligned}
\end{equation}
where the first term is the expected complete log-likelihood and the second term is the entropy of the proposal distribution $q$. It is guaranteed to be a lower bound of the observed log-likelihood $\log P(y_L | x_V)$ since their difference is the KL-divergence $D_{KL}(q(Y_U | x_v) \| p(Y_U | y_L, x_V))$ which is always positive.
\begin{equation}
\begin{aligned}
&\log P(y_L | x_V) - \mathcal{L}(q) \\
&= D_{KL} \big( q(Y_U | x_v) \| p(Y_U | y_L, x_V) \big) \geq 0. \label{eq:elbodiff}
\end{aligned}
\end{equation}
The EM algorithm is none other than an alternating maximization of ELBO with E-step and M-step. During E-steps, we fix the learned parameters $(\theta, K)$ of posterior distribution $p(Y_U|y_L,x_V;\theta, K)$ and update variational distribution $q(Y_U|x_V)$. During M-steps, we fix the proposal variational distribution $q(Y_U|x_V)$ and learn parameters $(\theta, K)$ that maximize the expected complete log-likelihood over all nodes w.r.t.\ the distribution of $q(Y_U|x_V)$.

\paragraph{E-Step}
In an E-step we fix parameters $(\theta, K)$ and update $q$ to maximize the ELBO $\mathcal{L}(q; \theta, K)$, which according to Equation~\eqref{eq:elbodiff} is equivalent to minimizing the KL-divergence $D_{KL} ( q(Y_U | x_v) \| p(Y_U | y_L, x_V) )$ since $\log P(y_L | x_V)$ does not depend on $q$.  So the update function during E-Step is
\begin{equation}
    q^*(Y_U|x_V) = \argmin_{q} D_{KL}(q(Y_U|x_V)\|p(Y_U|y_L,x_V)).
\end{equation}

The proposal distribution $q$ has two desiderata: it should closely approximate the true likelihood distribution of missing values $p(Y_U | y_L, x_V)$, and it should have a simple form such that the ELBO can easily be maximized. Here we restrict the proposal distribution $q$ to have a fully factorized form
\begin{equation}
q(Y_U|x_V) = \prod_{i \in U} q_i(Y_i|x_V), \label{eq:meanfield}
\end{equation}
so that the proposal distribution $q(Y_U | x_V)$ is a mean field approximation \citep{koller2009pgmbook} of $p(Y_U | y_L, x_V)$.

With the mean field approximation, there is a closed-form expression for updating the proposal distribution q when fixing the parameters $\theta$ and K: $\forall i \in U$, we have
\begin{equation}
\begin{split}
    \enspace q_i(Y_i|x_V) &\propto \exp \big( \log \phi_i(Y_i, x_v) \\
    &+ \sum_{j \in NB(i) \cap L} \log \phi_{i,j}(Y_i, y_j) \\
    &+ \sum_{k \in NB(i) \cap U} \sum_{Y_k} \log \phi_{i,k}(Y_i, Y_k) q_k(Y_k | x_V) \big).
\end{split}
\end{equation}
Since the mean field update is essentially a message-passing procedure, we could easily design a message passing layer using the above function to update the proposal distribution $q$.

\paragraph{M-Step}
In an M-step we fix the proposal distribution $q$ and update parameters  $(\theta, K)$ to maximize the ELBO $\mathcal{L}(q; \theta, K)$, which is equivalent to maximizing the expected complete log-likelihood $\mathds{E}_q[\log P(Y_U, y_L | x_V; \theta, K)]$ because the entropy term (second part in Equation~\eqref{eq:elbodef}) only depends on $q$. So the objective function during M-Step is
\begin{equation}
\theta^*, K^* = \argmax_{\theta, K}\mathds{E}_{q(Y_U|x_V)}[\log P(Y_U,y_L | x_V;\theta, K)].
\end{equation}

Computing the (expected) complete log-likelihood is also infeasible since the random variable grows exponentially. To make the learning procedure tractable, we apply the piecewise learning method, which we will discuss in the next section.

\subsubsection{Piecewise surrogate objective for M-Step} \label{sec: pw m-step}

The inference on MRF is intractable, and the computation of partition function $Z(x_V)$ in Equation~\eqref{eq:crffactor} is infeasible for graph data in practice. This hinders the learning process of our model. To overcome this, we utilize a tractable surrogate objective function to approximate the original likelihood. 

\paragraph{Star-shaped piecewise surrogate objective}
The piecewise training method provides an alternative local training approach \citep{sutton2009piecewise}. It breaks the whole graphical model down into tractable subgraphs and performs inference separately on each piece. The subgraphs can have overlaps among them and should have tractable structures for inference, e.g., trees. The factors contained in each subgraph are used to define its distribution. In our case, we want to model the interdependency between the central node and its neighbors. Thus, we star-wisely split the entire pairwise factor graph of the MRF: As shown in Figure~\ref{fig: PWEM}, for every node $i$ in the graph, we create a piece which groups this node and its neighbors as a subgraph. So the objective function approximates the original log-likelihood as
\begin{equation}\label{Eq:PW}
\begin{split}
    \log P(Y_V | x_V; \theta, K) & \approx \ell_{pw}(Y_V|x_V) \\
    &=\sum_{i \in V} \log P_i^*(Y_i, Y_{\mathcal{N}(i)} | x_V),
\end{split}
\end{equation}
where $\mathcal{N}(i)$ represents the neighbor nodes of central node $i$. This star-shaped piece is a tree structure rooted at the central node, and this piece is parametrized by the factors from the original MRF and pairwise factorized following Equation~\eqref{eq:crffactor}. Thus, the partition functions of the subgraphs can be easily computed using belief propagation~\citep{koller2009pgmbook}. With this approximation, we can design a message passing layer to estimate the surrogate objective function.

\paragraph{Factor redistribution}
For every node, we group this node and its connected neighborhood as a piece, which generates overlaps among different pieces as shown in Figure~\ref{fig: PWEM}. This means the factors that are contained in intersecting regions of different pieces will be overcounted. To avoid this issue, we redistribute the overlapping factor values in every subgraph. An even redistribution of factors is formulated as follows:
\begin{equation}\label{eq:average_redist}
\begin{split}
        \psi_i(Y_i, x_V) = \big( \phi_i(Y_i, x_V) \big)^{\frac{1}{d(i)+1}}\\ \psi_{j,k}(Y_j, Y_k)) = \big( \phi_{j,k}(Y_j, Y_k)) \big)^{\frac{1}{2}}.
\end{split}
\end{equation}

With this redistribution, the distribution of each piece $\bar{P}_i^*(Y_i, Y_{\mathcal{N}(i)} | x_V)$ becomes
\begin{equation}
\begin{split}
        &\bar{P}_i^*(Y_i, Y_{\mathcal{N}(i)} | x_V) \\
        &= \frac{1}{\bar{Z}_i^*(x_V)} \psi_i(Y_i, x_V) \prod_{j \in \mathcal{N}(i)} \psi_j(Y_j, x_V) \psi_{i,j}(Y_i, Y_j),
\end{split}
\end{equation}
so that the redistributed piecewise log likelihood $\bar{\ell}_{pw}(Y_V|x_V)$ of the whole model has the following form:
\begin{equation}
    \begin{split}
            \bar{\ell}_{pw}(Y_V | x_V) = &\sum_{i \in V} \log \phi_i(Y_i, x_V) \\
            &+ \sum_{(j,k) \in E, j < k} \log \phi_{j,k}(Y_j, Y_k) \\
             &- \sum_{i \in V} \log \bar{Z}_i^*(x_V).\\
    \end{split}
\end{equation}
Here we see that the factor overcounting issue is fixed. This piecewise likelihood reasonably approximates the original distribution $P(Y_V|x_V)$ since the only difference between Equation~\ref{eq:crffactor} is the partition function.

\paragraph{Edgewise scalar coefficient}
For our discussion so far, we only considered a single $c \times c$ compatibility matrix $K$ to parametrize pairwise factors $\phi_{j,k}(y_j, y_k)$ of every connected node pair, meaning that all connected node pairs share a common parametrization (denote $K(l,m) = K_{l,m}$): 
\begin{equation}
    \log \phi_{j,k}(y_j, y_k) = K(y_j, y_k).
\end{equation}
This representation can be restricting since it does not differentiate the edges in the graph. It is thus necessary to relax this constraint in order to equip our model with more representational power. However, if we parametrize every edge with a different $c\times c$ matrix, the model will easily suffer from overfitting. In order to balance these two aspects, we introduce an additional scalar parameter $\alpha_{j,k}$ for each edge $(j,k)$ that scales the shared compatibility matrix.
Thus, the parametrization becomes
\begin{equation}
    \log \phi_{j,k}(y_j, y_k) = \alpha_{j,k}\ K(y_j, y_k)
\end{equation}

and the star-shaped piecewise log-likelihood becomes
\begin{equation}
    \begin{split}
            \bar{\ell}_{pw}(Y_V | x_V) = &\sum_{i \in V} \log \phi_i(Y_i, x_V) \\
            &+ \sum_{(j,k) \in E, j < k} \alpha_{j,k}\ K(Y_j, Y_k) \\
             &- \sum_{i \in V} \log \bar{Z}_i^*(x_V).\\
    \end{split}
\end{equation}

\section{Experiment and Result}\label{sec:experiments}

In this section, we compare our model with state-of-the-art methods such as GCN \citep{kipf2016semi}, GAT \citep{velivckovic2017graph}, and GMNN \citep{qu2019gmnn} for semi-supervised node classification problems. We evaluate our model on a wide variety of datasets, including both assortative and disassortative graphs. We also conduct an ablation study to analyze different components of our model. 
We report average accuracy on $50$ runs of experiments for analysis.

\subsection{Dataset}
In this section, we introduce the characteristics of graph datasets including both assortative and disassortative graphs.

\paragraph{Citation networks} The citation network is a graph where nodes represent papers, edges depict the citation relationship between two papers, and node features are the bag of words of that paper. The node label is the topic of the corresponding paper. Cora, Citeseer and PubMed are common benchmark datasets \citep{namata2012query} \citep{sen2008collective} of this type. We employ the same configuration as in \citet{yang2016revisiting}, which randomly picks $20$ samples out of every class as training nodes, and randomly selects $500$ and $1000$ samples for validation and test, respectively. 

\paragraph{Wikipedia networks} Wikipedia networks are graphs where nodes stand for websites, and edges represent the mutual links of two pages. Node features are representative nouns of that page, and nodes are classified according to their average monthly traffic. Chameleon and Squirrel \citep{rozemberczki2019multi} are both examples of such networks. We split the graph to the train, validation, and test according to the following ratio: $20\%$, $20\%$, and $60\%$.

\paragraph{Actor co-occurrence networks} Actor co-occurrence network is a graph extracted from film-director-actor-writer network \citep{tang2009social}. The Actor dataset is one of the graph where nodes represent actors and edges indicate co-occurrence relationship in the same website between two actors. The node features are representative keywords of an actor. We split the graph in the same manner as for Wikipedia networks.

\begin{table}[ht]
        \centering
        \caption{Statistics of common datasets. $\beta$ is defined in Section~\ref{sec:graph_homo} and indicates the graph homophily. Higher $\beta$ means the connected nodes in the graph tend to agree with each other and have the same labels.}
        \label{tab:dataset}
        \begin{tabular}{ccccccccc}
        \hline
   Dataset &  Nodes/Edges &  Features/Classes & $\beta$ \\
\hline 
Cora   & 2708 / 5429 & 1433 / 7 & 0.83 \\
Citeseer   & 3327 / 4732 & 3703 / 6 &0.71 \\
Pubmed  & 19717 / 44338 & 500 / 3 & 0.79 \\
Chameleon & 2277 / 36101 & 2325 / 5 & 0.25 \\
Squirrel & 5201 / 217073 & 2089 / 5 &0.22\\
Actor & 7600 / 33544 & 931 / 5 &0.24\\
\hline
        \end{tabular}
        
\end{table}

\subsection{Classification on graphs}
To evaluate the effectiveness of our method, we compare our model with several baseline methods on semi-supervised node classification tasks. We use the Cora, Citeseer, and Pubmed datasets for comparison. For our EPFGNN model, we utilize GCN as the graph neural network backbone to learn the representation of input--output factors given input features. We use the same hyperparameter setting as described in \citep{kipf2016semi}. We employ a compatibility matrix to model the interdependency among node labels and extend the pairwise factors with a scalar coefficient for every edge. The test accuracies of different models on different datasets are collected in Table~\ref{tab:results}.

\begin{table}[ht]
    \centering
    \caption{Mean classification accuracy of the proposed EPFGNN model compared with baseline methods on Cora, Citeseer and Pubmed datasets.}
    \label{tab:results}
    \begin{tabular}{cccc}
 \hline
Model  & Cora & Citeseer & Pubmed  \\
 \hline
GCN & 81.56 & 70.37 & 78.69  \\  
GAT & 82.08 & 71.44 & 77.52\\  
GMNN & 82.05 & 70.53 & 79.38 \\
EPFGNN & \textbf{83.54}  & \textbf{73.13 } & \textbf{80.15} \\

 \hline
 \end{tabular}
\end{table}

 As shown in Table~\ref{tab:results}, our method outperforms other state-of-the-art baselines, especially on the Citeseer dataset, where we observe a performance improvement for over $2\%$. These results support our analysis that modeling the direct interdependency among node labels will improve performance. Also, a more significant performance increase is observed on the Citeseer dataset, which has the lowest graph homophily $\beta$ among three datasets. This shows the advantage of the EPFGNN representation: contrary to other baselines methods, it goes beyond feature smoothing.

\subsection{Analysis of Graph Homophily}\label{sec:graph_homo}

The citation networks are assortative graphs since the connected nodes tend to have the same label. In contrast, the Wikipedia networks and actor co-occurrence networks are disassortative graphs where the assumption of graph homophily no longer holds. To quantify the degree of graph homophily, we use the measure $\beta$ from \citet{pei2020geom} defined as follows:
\begin{equation}
    \beta = \frac{1}{V} \sum_{v\in V} \frac{\textit{Number of v's neighbors with same label}}{\textit{Number of v's neighbors}}.
\end{equation}

The measure $\beta$ ranges between 0 and 1, where assortative graphs like Cora  have high $\beta$ values above $0.5$ while disassortative ones like Chameleon have low  $\beta$ values below $0.5$. $\beta$ value for various graph datasets are provided in Table~\ref{tab:dataset}.

We further experiment with disassortative graph datasets Chameleon, Squirrel, and Actor. We report the test accuracy results in Table~\ref{tab:disattortaive}.

 \begin{table}[ht]
    \centering
    \caption{Mean classification accuracy on disassortative graphs. We compare our model with GCN and GMNN on Chameleon, Squirrel and Actor datasets which have small $\beta$ values.}
    \label{tab:disattortaive}
    \begin{tabular}{cccc}
         \hline
         Model & Chameleon & Squirrel & Actor \\
         \hline
         GCN & 34.66  & 24.56  & 26.85 \\ 
         GMNN &34.69 & 24.77 &  \textbf{27.04} \\ 
         EPFGNN & \textbf{35.31} & \textbf{25.12} & 26.66\\
         \hline 
    \end{tabular}
\end{table}

As shown in Table~\ref{tab:disattortaive}, to a certain degree, our model managed to adapt to disassortative graphs. Since we have a shared compatibility matrix and every edge has one additional degree of freedom from the edgewise scalar coefficient, the explicit pairwise modeling in EPFGNN enables the model to flexibly aggregate the neighbor node label information when the neighboring nodes tend to disagree.

\subsection{Ablation Study}\label{sec:ablation study}

The proposed EPFGNN model learns the node representation and aggregates estimated label information from connected neighbor nodes. The key components of this model are the GNN backbone and the parametrization of pairwise factors. In this section, we compare different model variants to examine the contribution of each component.   

\paragraph{GNN-Backbone}
The role of the GNN backbone in the EPFGNN model is to model the influence of input features on output labeling. It parametrizes the input--output factors in the MRF. To find an appropriate choice, we consider two commonly used GNN models, GCN and GAT, as backbones for node representation learning, and evaluate their effectiveness by conducting experiments on three standard datasets Cora, Citeseer, and Pubmed. 

\begin{table}[ht]
    \centering
    \caption{Comparison of classification accuracy with EPFGNN using GCN and GAT backbones.}
    \label{tab:ab_backbone}
    \begin{tabular}{cccc}
         \hline
         GNN backbone & Cora & Citeseer & Pubmed \\
         \hline
         GCN & \textbf{83.24}  & \textbf{72.35 }  & \textbf{79.61} \\ 
         GAT & 82.32  & 71.99  &79.23\\
         \hline
    \end{tabular}
\end{table}

In Table~\ref{tab:ab_backbone}, we observe that GCN outperforms GAT on all three datasets. These results indicate that GAT, which has more parameters compared to GCN, is likely to suffer from overfitting as GNN backbone for EPFGNN.

\paragraph{Redistribution}
The star-shaped split of the whole graph will cause overlaps among pieces and result in the overcounting of factors when estimating the piecewise likelihood. To address this problem, we redistribute the factor values as described in Equation~\eqref{eq:average_redist}. We refer to this distribution method as average redistribution. To verify whether this redistribution is reasonable, we compare it with a different redistribution method referred to as center redistribution.  For center redistribution, we assign each input--output factor $\phi_i(Y_i, x_V)$ entirely to the subgraph where $i$ is the central node. This yields a different estimation of the piecewise likelihood. We compare these two redistribution methods on the three standard datasets Cora, Citeseer, and Pubmed and collect the classification accuracies in Table~\ref{tab:ab_redistribution}.

\begin{table}[ht]
    \centering
    \caption{Mean classification of EPFGNN with different redistribution methods. For all experiments, we use GCN as GNN backbone and do not use the additional edgewise coefficient.}
    \label{tab:ab_redistribution}
    \begin{tabular}{cccc}
         \hline
         Factor redistribution & Cora & Citeseer & Pubmed \\
         \hline
         average& \textbf{83.24}  & 72.35  & \textbf{79.61} \\ 
         center & 82.42  & \textbf{72.46}& 78.67\\
         \hline
    \end{tabular}
\end{table}

The results in Table~\ref{tab:ab_redistribution} show that average redistribution tends to yield better results, possibly owing to the fact that the input--output factors provided by the GNN backbone is accessible in all related subgraphs. It remains an interesting future work to analyze other possible redistribution schemes.

\paragraph{Edgewise scaling coefficient}
To understand the effect of introducing the edgewise scaling coefficient, we compare it with two alternative settings. We name the setting that only has shared parametrization for all output--output factors as EPFGNN w/o coefficient and name the setting with both shared compatibility matrix and shared scalar coefficient for output--output factors on every edge as EPFGNN using layer coefficient. We refer to the original setting, which has a shared compatibility matrix and edgewise scalar coefficients, as EPFGNN using edge coefficient. Their comparison is summarized in Table~\ref{tab:ab_edgerezero}.

\begin{table}[ht]
    \centering
    \caption{Mean accuracy of EPFGNN with different output--output factor parametrization settings. For all experiments, we use GCN backbone and average redistribution.}
    \label{tab:ab_edgerezero}
    \begin{tabular}{cccc}
         \hline
         Scaling factor & Cora & Citeseer & Pubmed \\
         \hline
         w/o coefficient & 83.24  & 72.35  & 79.61 \\ 
         layer coefficient & 83.03 & 72.33& 78.95\\
         edge coefficient &\textbf{83.54}  & \textbf{73.13 } & \textbf{80.15} \\ 
         \hline
    \end{tabular}
\end{table}

The results in Table~\ref{tab:ab_edgerezero} show that the edgewise coefficient setting outperforms other alternative settings. Thus we can conclude that the additional flexibility provided by the edgewise scaling coefficient is beneficial and can effectively improve the representational power of the EPFGNN model.

\section{Conclusion}\label{sec:conclusion}

This paper proposes the novel EPFGNN framework in which explicit pairwise factors are defined to model direct dependency between connected node pairs. In this way, the model can simultaneously aggregate node input features and label information from neighbor nodes. By introducing a shared compatibility matrix and edgewise scaling coefficients, we are able to effectively provide a flexible representation while avoiding overfitting. With the application of the EM algorithm and the surrogate piecewise objective in the M-step, we manage to leverage information from unobserved nodes and make the learning procedure tractable. We validate our analysis and model design with a series of experiments, which shows that the EPFGNN framework can effectively handle semi-supervised node classification problems for various graph datasets.

\begin{acknowledgements} 
    This work was supported by the Munich Center for Machine Learning. We would like to thank Stephan Günnemann for helpful discussions.
\end{acknowledgements}

\bibliography{wang_745}
\end{document}